\documentclass{article}
\usepackage{spconf,amsmath,graphicx,epstopdf,amsmath,subcaption,nccmath,multirow,hyperref}

\title{FDFlowNet: \\Fast Optical Flow Estimation using a Deep Lightweight Network}
\name{Lingtong Kong$^{1}$ \quad Jie Yang$^{1,2}$}
\address{${}^{1}$Institute of Image Processing and Pattern Recognition, Shanghai Jiao Tong University, China\\
${}^{2}$Institute of Medical Robotics, Shanghai Jiao Tong University, China}
\begin{document}
%
\maketitle
\begin{abstract}
Significant progress has been made for estimating optical flow using deep neural networks. Advanced deep models achieve accurate flow estimation often with a considerable computation complexity and time-consuming training processes. In this work, we present a
lightweight yet effective model for {\it real-time} optical flow estimation, termed FDFlowNet (fast deep flownet). We achieve better or similar accuracy on the challenging KITTI and Sintel benchmarks while being about 2 times faster than PWC-Net. This is achieved by a carefully-designed structure and newly proposed components. We first introduce an {\it U-shape} network for constructing multi-scale feature which benefits upper levels with global receptive field compared with pyramid network. In each scale, a {\it partial fully connected} structure with dilated convolution is proposed for flow estimation that obtains a good balance among speed, accuracy and number of parameters compared with sequential connected and dense connected structures. Experiments demonstrate that our model achieves state-of-the-art performance while being fast and lightweight.
\end{abstract}
\section{Introduction}
Optical flow estimation is a fundamental problem in computer vision which plays an important role in many vision applications such as action recognition~\cite{NIPS2014_5353}, video understanding~\cite{Fan_2018_CVPR} and self-driving
cars~\cite{8658397,Kong2019ICONIP}. With decades of research, approaches based on energy minimization~\cite{Horn:1981} perform best among all methods.  Optimization of an energy function in a coarse-to-fine manner is typically computationally heavy, hampering it from real-time applications.

Recently deep convolutional neural networks (CNNs) have dominated many fields of computer vision for being end-to-end trainable and powerful feature extraction ability. With advanced parallel computing equipments, CNNs can be inferred in real time. ~\cite{Fischer2015FlowNetLO} firstly apply CNNs to optical flow estimation and put forward two models namely FlowNetS and FlowNetC. Although the accuracy is close to  state-of-the-art energy minimization approaches, FlowNet's inference speed is orders of magnitude faster. By stacking multiple diverse Networks (FlowNetC, FlowNetS) and exploring different training schedules on multiple datasets, FlowNet2~\cite{Ilg2017FlowNet2E} improves both estimation accuracy and inference speed considerably. A drawback of FlowNet2 is
the model size (over 160M parameters), making it difficult for deployment to embedded devices. SPyNet~\cite{Ranjan2017OpticalFE} adopts a spatial pyramid network and warps the second image to the first one to estimate the residual flow in a coarse-to-fine manner. Compared with FlowNet, SPyNet has a much smaller-sized model (1.2M parameters) with the price of decreasing performance and running speed.

Compared with these previous works, recent works of
PWC-Net~\cite{Sun_2018_CVPR} and LiteFlowNet~\cite{Hui_2018_CVPR} use a similar design paradigm which can both increase performance and reduce model sizes. First, both models use feature pyramids to replace image pyramids for a better representation. Warping and correlation operations are also adopted at each pyramid level to reduce error as soon as possible. Both networks are top performing on several optical flow benchmarks, which confirms the superiority of these structures in optical flow estimation. Our work is mainly inspired by the success of PWC-Net and LiteFlowNet, but we go into
the overall architecture and explore the design of several key elements to reduce parameters and computation while obtaining high performance. In summary our contributions are:

$\bullet$ We propose a compact and effecient U-shape network as a improvement of pyramid network for optical flow estimation that can efficiently fuse multi-scale information while saving computing resources.

$\bullet$ We present a new partial fully connected structure of flow estimation network as an integration of existed dense connected and sequential connected structures. It provides a tradeoff among model size, computation cost and network performance.

$\bullet$ We show a new fast and lightweight deep neural network for optical flow estimation named FDFlowNet that achieves state-of-the-art performance on the challenging KITTI and Sintel benchmarks. It runs about 2 times faster than PWC-Net~\cite{Sun_2018_CVPR} and about 3.2 times faster than LiteFlowNet~\cite{Hui_2018_CVPR} on Sintel resolution images.

\section{Method}
Our goal is to estimated the dense optical flow field $F$ given two time adjacent frames $I_1$ and $I_2$. Working in a coarse-to-fine manner, FDFlowNet consists of one U-shape network for constructing multi-scale feature and five flow estimation networks in different levels.

\subsection{U-shape Network}

\begin{figure}[t]
	\centering
	\includegraphics[scale=0.38]{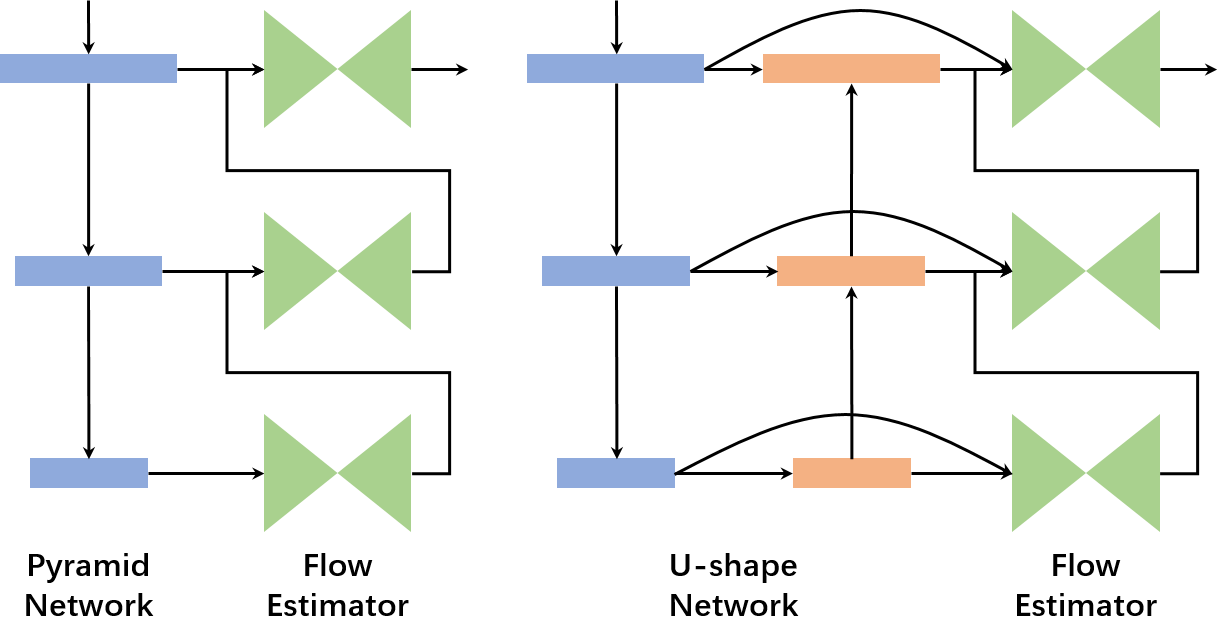}
	\caption{The left is traditional pyramid network for optical flow estimation adopted in PWC-Net and LiteFlowNet, the right is our proposed U-shape network. Only three levels are shown for brevity.}
	\label{fig:1}
\end{figure}

The U-shape network architecture plays an important role in FDFlowNet as shown in Fig.~\ref{fig:1}.
Two frames $I_1$ and $I_2$ are imported into the same pyramid network from level 1 to level 6. When passing through level $k$, it generates two image feature $f_1^k$ and $f_2^k$. Different from previous works of PWC-Net~\cite{Sun_2018_CVPR} and LiteFlowNet~\cite{Hui_2018_CVPR} that directly use pyramid feature, we build a symmetrical part for fusing low-level and high-level information that forms the U-shape network. 

A drawback of pyramid network is that the higher resolution where flow estimator locates, the less semantic information corresponding pyramid feature contains. So it takes more convolution layers for the flow estimator to rebuild it. On the other hand, this semantic information is similar among different resolutions, which means that computation on multiple levels can be redudant. 

To keep local property for dense matching, original pyramid feature $f_1^k$ and $f_2^k$ are used to calculate cost volume. Differently, semantic information is now offered by the fused feature ${\hat f}_1^k$ in U-shape network instead of $f_1^k$. From the first to the sixth level, the number of feature maps are 32, 64, 80, 96, 112 and 128. Transposed convolution which outputs half of input channels and one following convolution layer for feature fusing are employed in each level that constitute the U-shape network. Note that number of channels in pyramid feature $f_1^k$ is the same as that in fused feature ${\hat f}_1^k$.

\begin{figure}[t]
	\centering
	\includegraphics[scale=0.38]{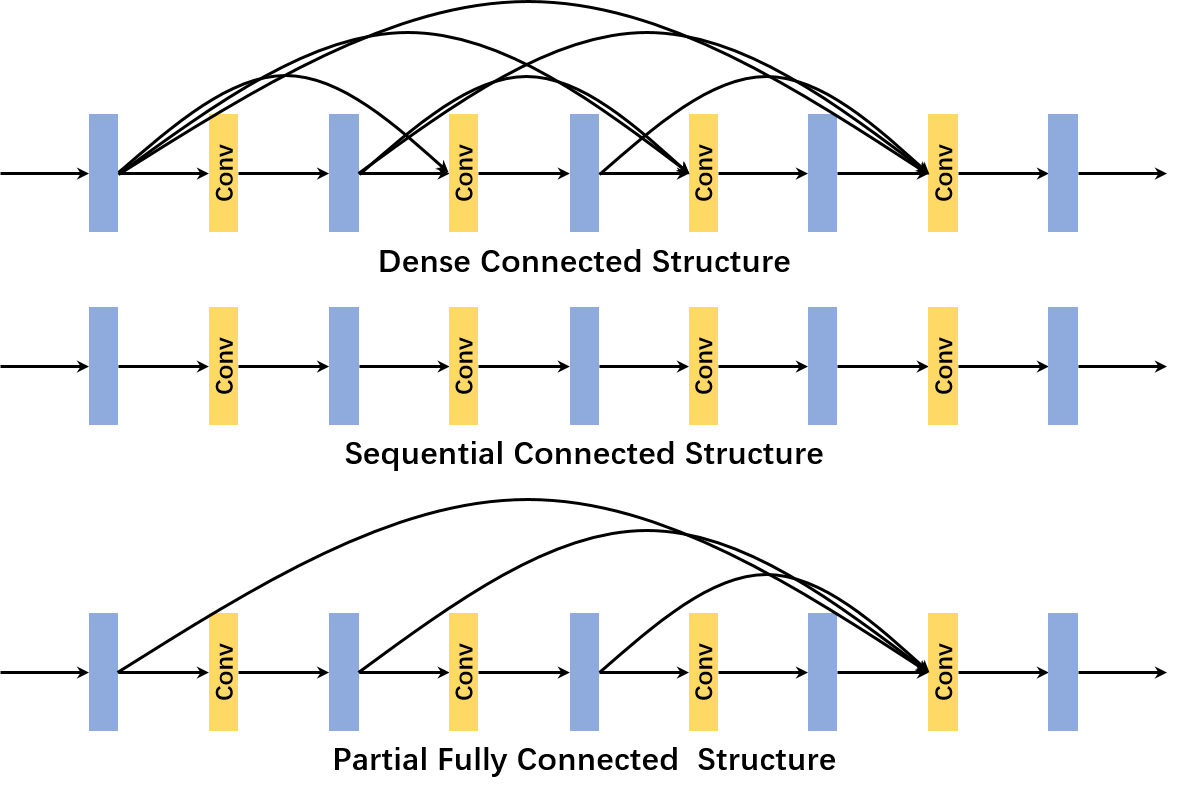}
	\caption{Comparison of three different modes of flow estimator. Partial fully connected structure used in our FDFlowNet provides a balance and tradeoff between the other two types in model size, computation cost and network performance.}
	\label{fig:2}
\end{figure}

\subsection{Warping, Correlation and Aggregation}
One challenge in optical flow estimation is the large displacement of objects in two neighboring frames. To solve the problem, an effective method is to use the initially estimated flow to warp~\cite{10.1007/978-3-540-24673-2_3} the second feature map. Given two feature map $f_1^k$ and $f_2^k$, we use upsampled flow in level $(k+1)$ to resample $f_2^k$ with bilinear interpolation that generates the warped second pyramid feature $f_{warp}^k$.

To get better correspondence comparison between two feature of $f_1^k$ and $f_{warp}^k$, we employ the correlation layer in ~\cite{Fischer2015FlowNetLO} to build a 3D cost volume which can be formulated as
\begin{displaymath}
c^k({\bf x}, {\bf d}) = {f_1^k({\bf x})} \cdot {f}_{warp}^k({\bf x}+{\bf d}) / N.
\end{displaymath}
$\bf x$ and $\bf d$ represent spatial and offset two-dimension coordinates respectively. $N$ is the number of channels in cost volume. We set search radius to 4 in all levels as~\cite{Sun_2018_CVPR}, and we have found that adding a convolution layer after rigid cost volume for feature aggregation can improve performance. The aggregated cost volume $c_{aggr}^k$ has 126 channels in each scale.

\begin{table*}[t]
\footnotesize
\renewcommand\arraystretch{1.1}
\begin{center}
\centering
\resizebox{1.0\textwidth}{!}{
	\begin{tabular}{r|cc|cc|ccc|ccc}
		Method & \multicolumn{2}{|c|}{Sintel Clean} & \multicolumn{2}{|c|}{Sintel Final} & \multicolumn{3}{|c|}{KITTI12} & \multicolumn{3}{|c  }{KITTI15}\\
		& train & test & train & test & train & test & test(Fl-Noc) & train & train(Fl-all) & test(Fl-all)\\
		\hline
		FlowNetC \cite{Fischer2015FlowNetLO} & 4.31 & 7.28 & 5.87 & 8.81 & 9.35 & - & - & - & - & - \\
		FlowNetC-ft & (3.78) & 6.85 & (5.28) & 8.51 & 8.79 & - & - & - & - & - \\
		SPyNet \cite{Ranjan2017OpticalFE} & 4.12 & 6.69 & 5.57 & 8.43 & 9.12 & - & - & - & - & - \\
		SPyNet-ft & (3.17) & 6.64 & (4.32) & 8.36 & (4.13) & 4.7 & 12.31\% & - & - & 35.07\% \\
		FlowNet2 \cite{Ilg2017FlowNet2E} & 2.02 & 3.96 & 3.14 & 6.02 & 4.09 & - & - & 10.06 & 30.37\% & - \\
		FlowNet2-ft & (1.45) & 4.16 & (2.01) & 5.74 & (1.28) & 1.8 & 4.82\% & (2.30) & (8.61\%) & 11.48\% \\
		PWC-Net \cite{Sun_2018_CVPR} & 2.55 & - & 3.93 & - & 4.14 & - & - & 10.35 & 33.67\% & - \\
		PWC-Net-ft & (2.02) & 4.39 & (2.08) & {\bf 5.04} & (1.45) & 1.7 & 4.22\% & (2.16) & (9.80\%) & 9.60\% \\
		PWC-Net-small \cite{Sun_2018_CVPR} & 2.83 & - & 4.08 & - & - & - & - & - & - & - \\
		PWC-Net-small-ft & (2.27) & 5.05 & (2.45) & 5.32  & - & - & - & - & - & - \\
		LiteFlowNet\cite{Hui_2018_CVPR} & 2.48 & - & 4.04 & - & 4.00 & - & - & 10.39 & 28.50\% & - \\
		LiteFlowNet-ft & {\bf (1.35)} & 4.54 & {\bf (1.78)} & 5.38 & {\bf (1.05)} & 1.6 & 3.27\% & (1.62) & {\bf (5.58\%)} & {\bf 9.38\%} \\
		FDFlowNet & 2.60 & - & 4.12 & - & 4.13 & - & - & 10.75 & 29.59\% & - \\
		FDFlowNet-ft & (1.80) & {\bf 3.71} & (1.93) & 5.11 & (1.09) & {\bf 1.5} & {\bf 3.19\%} & {\bf (1.56)} & (6.36\%) & {\bf 9.38\%} \\
\end{tabular}}
\end{center}
\caption{Performance comparison on challenging benchmarks. Default metric is end point error (EPE). KITTI benckmarks are measured by `Fl-all' (Fl-all: Percentage of outliers averaged over all pixels. Inliers are defined as EPE$<$3 pixels or $<$5\%).}
\label{tab:1}
\end{table*}

\begin{figure*}[ht]
	\setlength{\abovecaptionskip}{2pt}
	\centering
	\begin{minipage}[b]{0.24\textwidth}
		\includegraphics[scale=0.116]{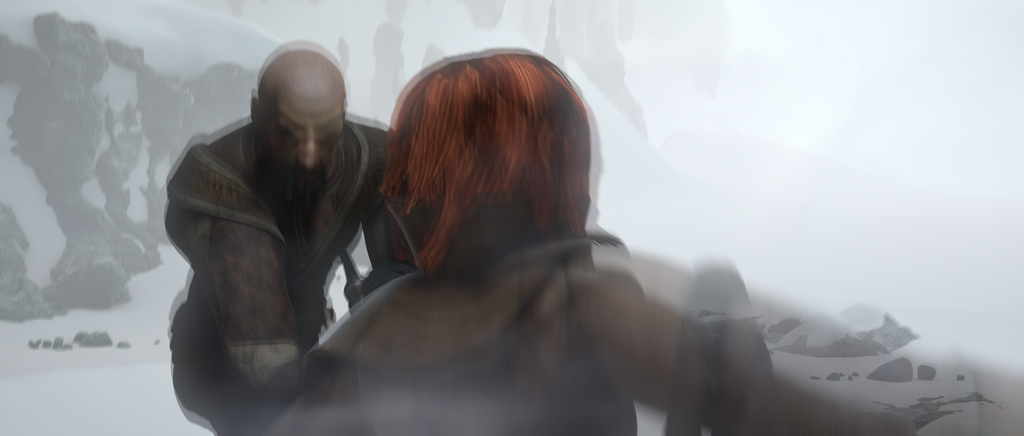}
	\end{minipage}
	\begin{minipage}[b]{0.24\textwidth}
		\includegraphics[scale=0.116]{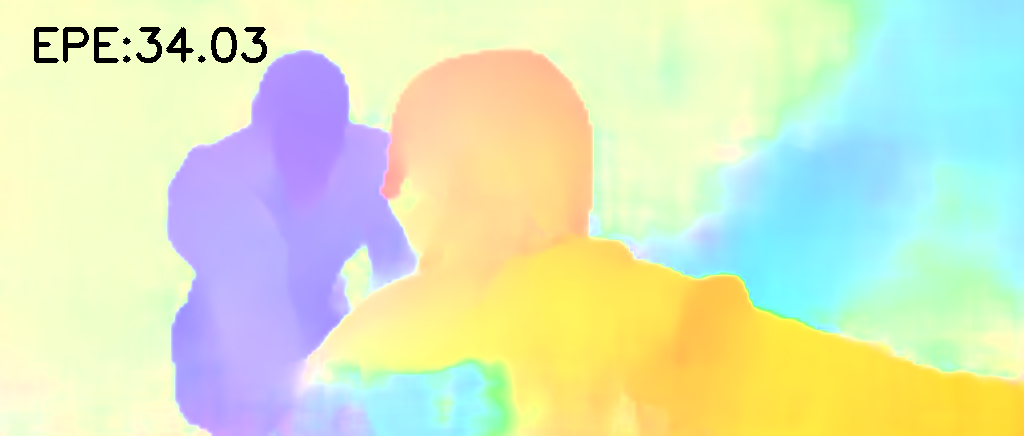}
	\end{minipage}
	\begin{minipage}[b]{0.24\textwidth}
		\includegraphics[scale=0.116]{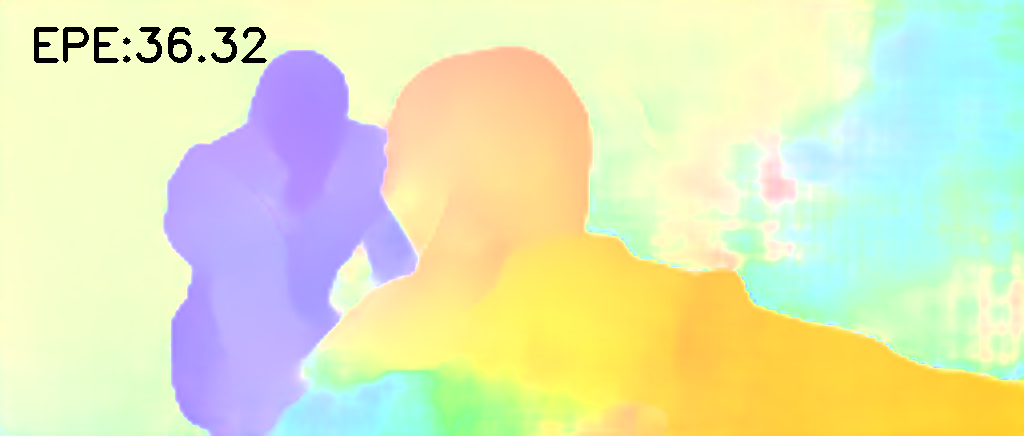}
	\end{minipage}
	\begin{minipage}[b]{0.24\textwidth}
		\includegraphics[scale=0.116]{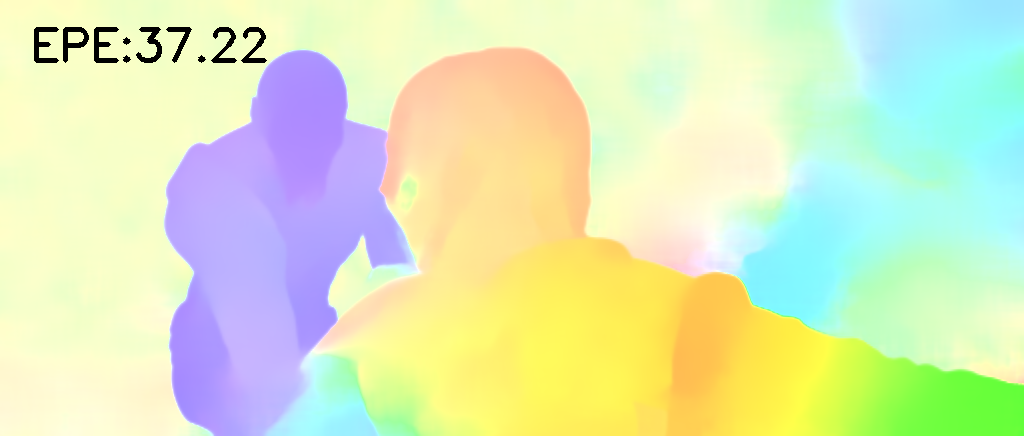}
	\end{minipage}
	
	\begin{minipage}[b]{0.24\textwidth}
		\includegraphics[scale=0.116]{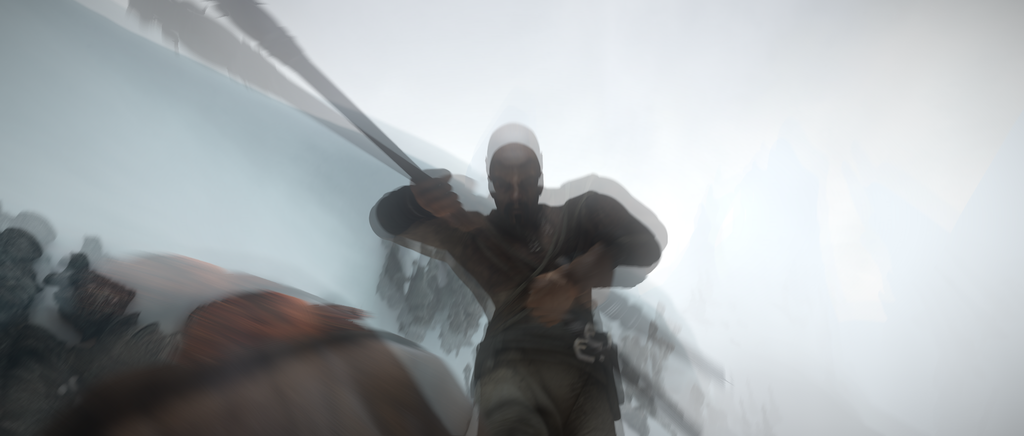}
		\subcaption*{Fused Image}
	\end{minipage}
	\begin{minipage}[b]{0.24\textwidth}
		\includegraphics[scale=0.116]{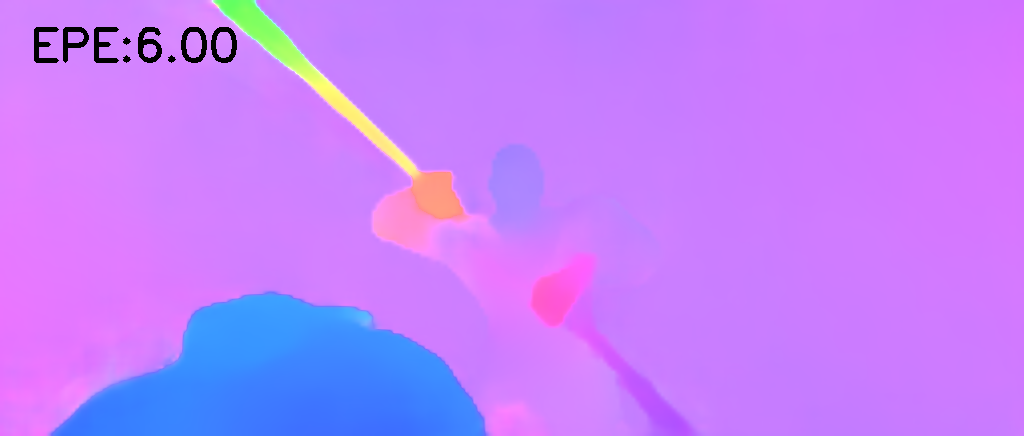}
		\subcaption*{FDFlowNet}
	\end{minipage}
	\begin{minipage}[b]{0.24\textwidth}
		\includegraphics[scale=0.116]{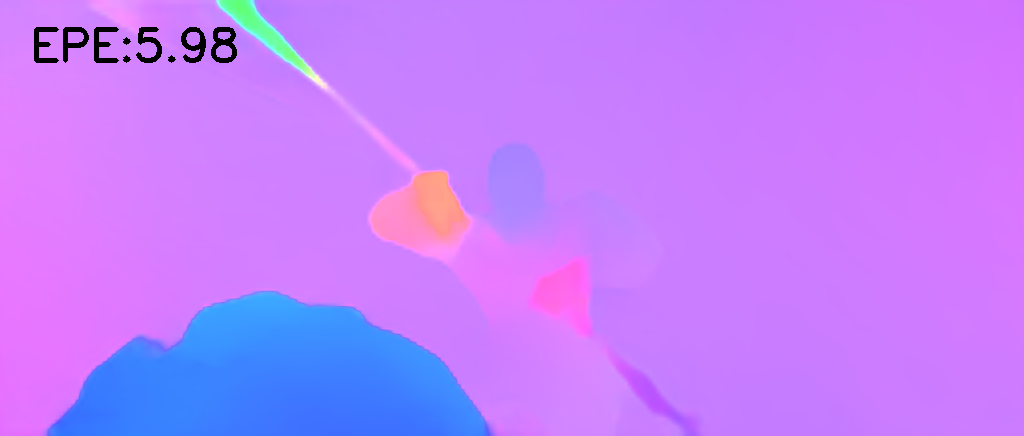}
		\subcaption*{PWC-Net}
	\end{minipage}
	\begin{minipage}[b]{0.24\textwidth}
		\includegraphics[scale=0.116]{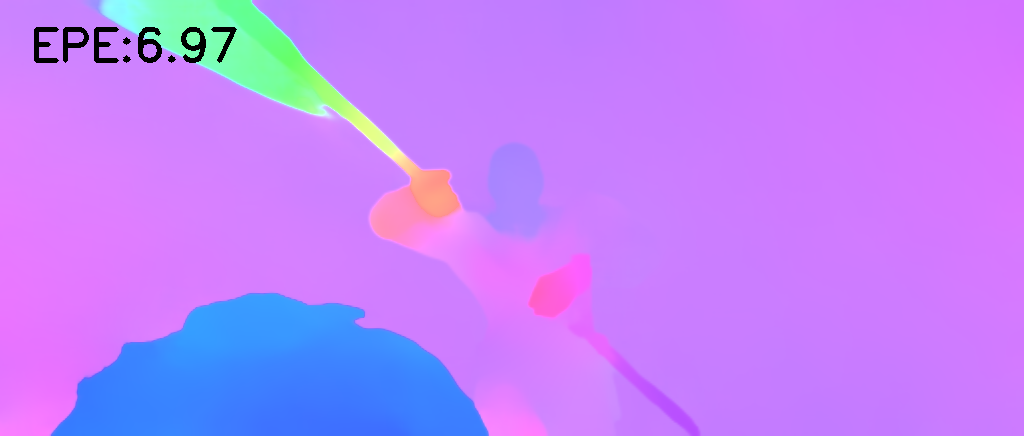}
		\subcaption*{LiteFlowNet}
	\end{minipage}
	\caption{Visualized optical flow field on the Sinel Final test set.}
	\label{fig:3}
\end{figure*}

\subsection{Partial Fully Connected Flow Estimator}
As mention above, fused feature of the first image ${\hat f}_1^k$, aggregated cost volume $c_{aggr}^k$ and upsampled flow in previous level ${\sf up_2}(F^{k+1})$ are used for estimating flow field $F^k$. The work in~\cite{Sun_2018_CVPR} has tested two types of flow estimation network: sequential connected structure and dense connected structure. Their experiments have shown that dense connected structure can get a better result after fine-tuning. However, it takes more parameters and computational cost. ~\cite{Hui_2018_CVPR} use a sequential connected structure for flow estimation and also get relatively good results.

Inspired by the above two connection manners, we proposed a new partial fully connected structure that provides a tradeoff between the other two types in model size, computation cost and network performance. These three structures are depicted in Fig.~\ref{fig:2}. Output feature maps of flow estimation network in FDFlowNet are 128, 128, 128, 96, 64, 32 and 2 where the dense connection is only adopted in the second last convolution. Follwing ~\cite{8682229, 8451790}, we change the convolution layers which output 64 and 32 channels to dilated convolution with dilation rate 2 for enlarging receptive field.

\section{Experiments}
\subsection{Training Details}
We use the same multi-scale training loss as PWC-Net~\cite{Sun_2018_CVPR}. All experiments are conducted on one NVIDIA 1080Ti GPU. We implement codes in PyTorch and adopt Adam~\cite{article} optimizer. Weight decay is set to $1e-4$ for regularization. For fair comparison, we first train FDFlowNet on FlyingChairs~\cite{Fischer2015FlowNetLO} dataset and then fine-tune it on FlyingThings3D~\cite{MIFDB16} dataset. $S_{short}$ and $S_{fine}$~\cite{Ilg2017FlowNet2E} schedules are employed for the two stage training respectively. We use the same data augmentation method including mirror, translate, zoom, rotate, squeeze, color jitter and random noise. We call the model {\bf FDFlowNet} after this sequential training procedure.

\begin{figure*}[ht]
	\setlength{\abovecaptionskip}{2pt}
	\centering
	\begin{minipage}[b]{0.24\textwidth}
		\includegraphics[scale=0.098]{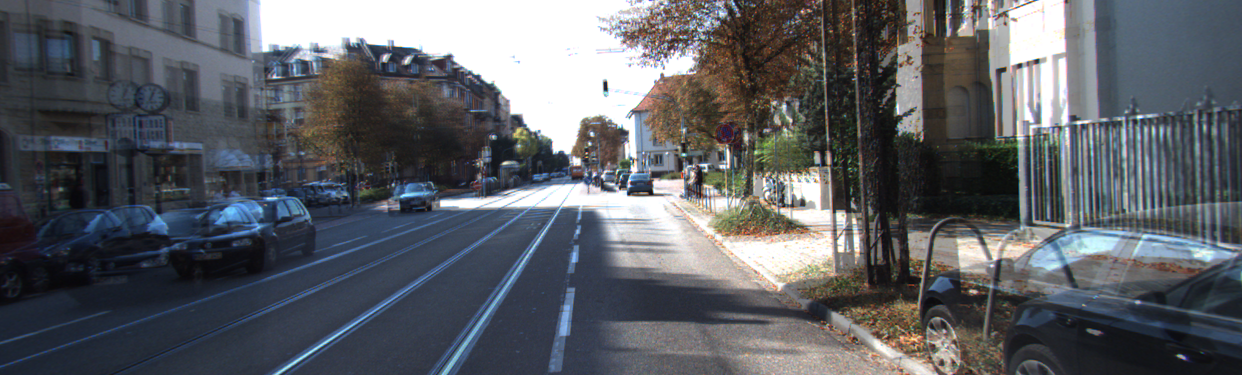}
	\end{minipage}
	\begin{minipage}[b]{0.24\textwidth}
		\includegraphics[scale=0.098]{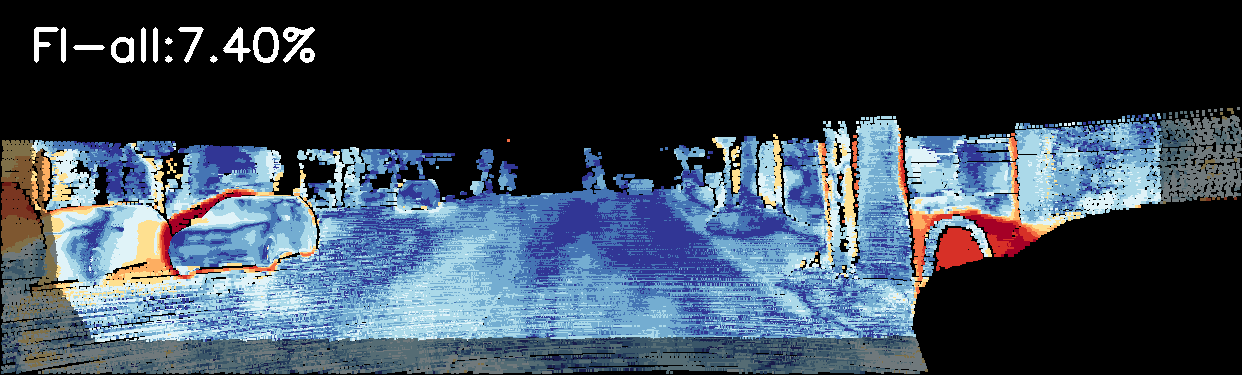}
	\end{minipage}
	\begin{minipage}[b]{0.24\textwidth}
		\includegraphics[scale=0.098]{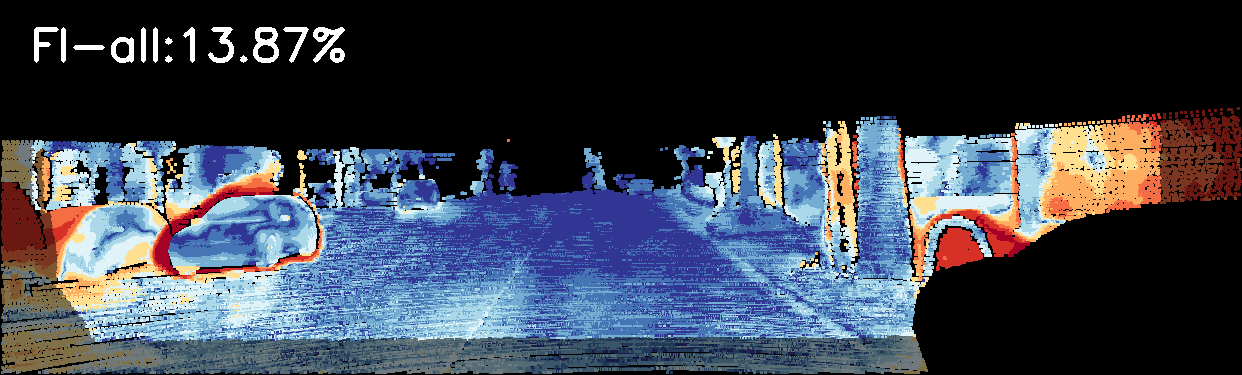}
	\end{minipage}
	\begin{minipage}[b]{0.24\textwidth}
		\includegraphics[scale=0.098]{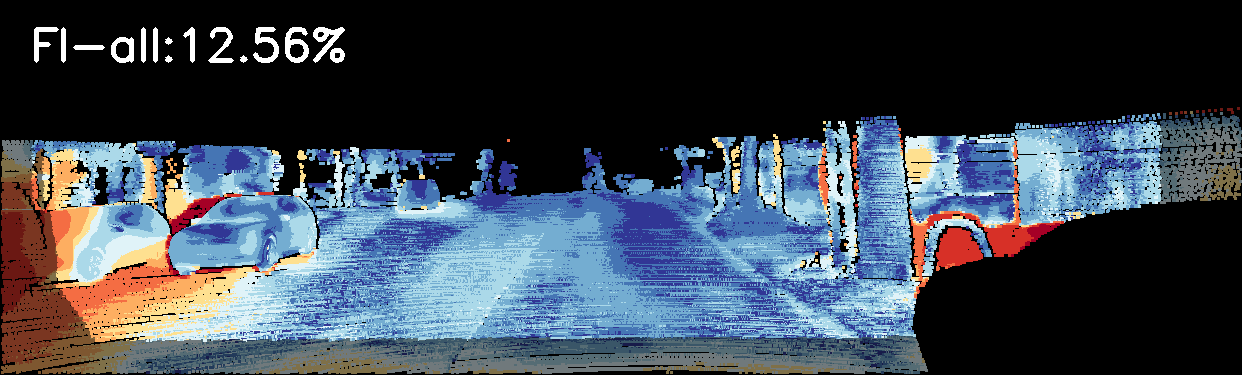}
	\end{minipage}

	\begin{minipage}[b]{0.24\textwidth}
		\includegraphics[scale=0.098]{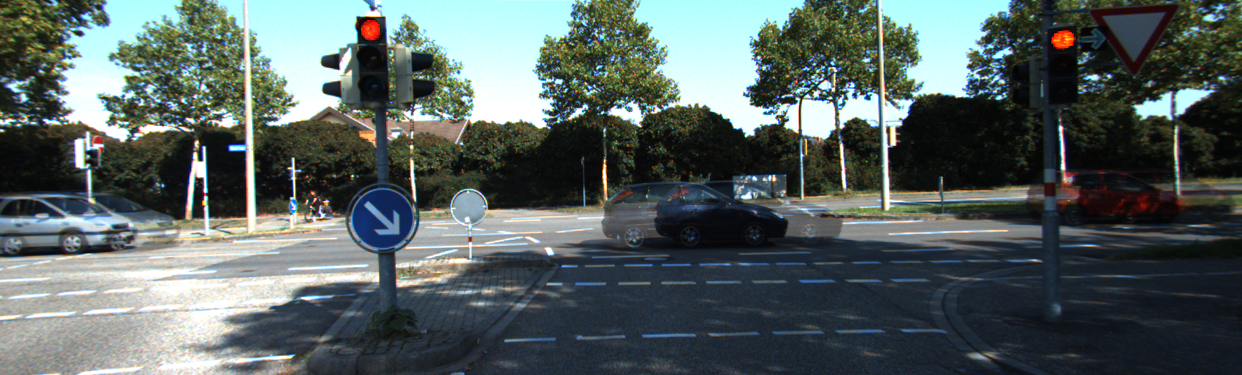}
	\end{minipage}
	\begin{minipage}[b]{0.24\textwidth}
		\includegraphics[scale=0.098]{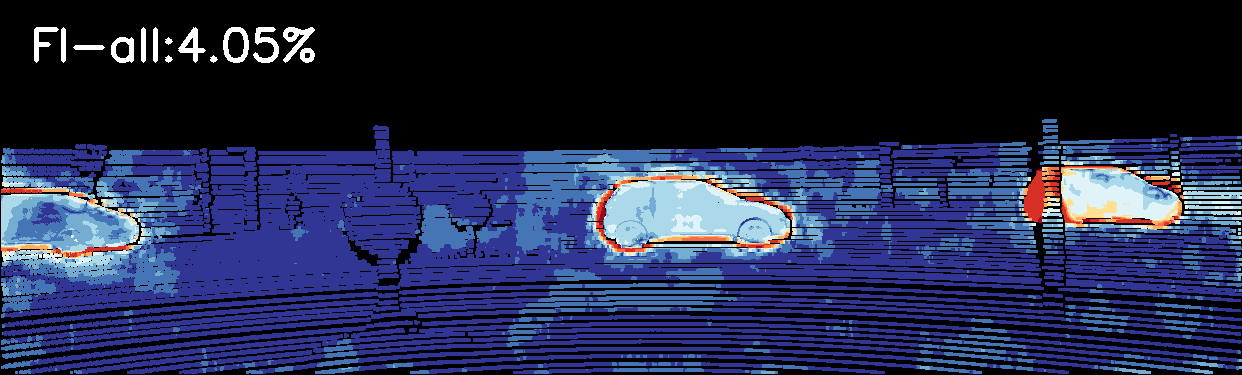}
	\end{minipage}
	\begin{minipage}[b]{0.24\textwidth}
		\includegraphics[scale=0.098]{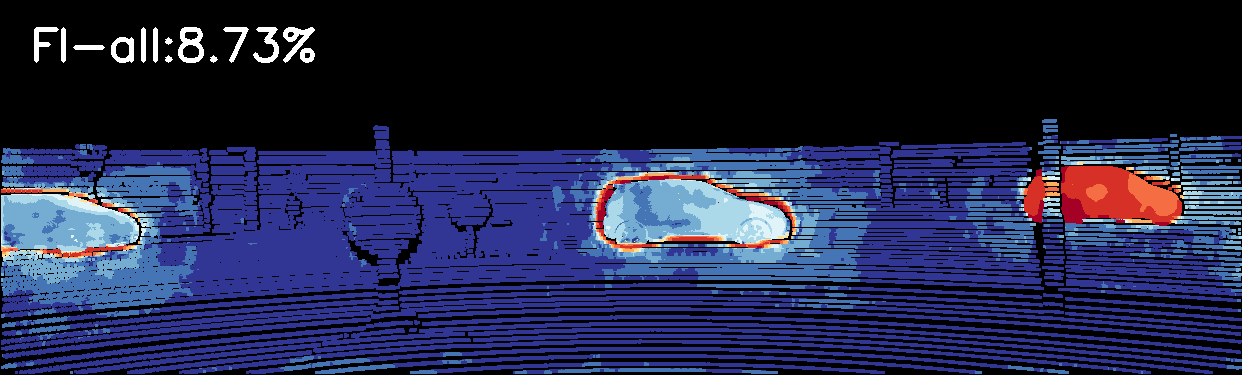}
	\end{minipage}
	\begin{minipage}[b]{0.24\textwidth}
		\includegraphics[scale=0.098]{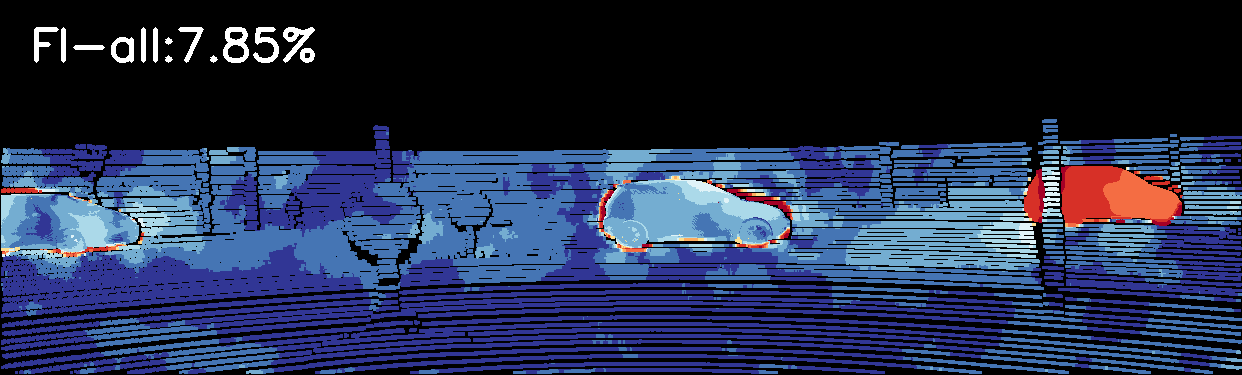}
	\end{minipage}

	\begin{minipage}[b]{0.24\textwidth}
		\includegraphics[scale=0.098]{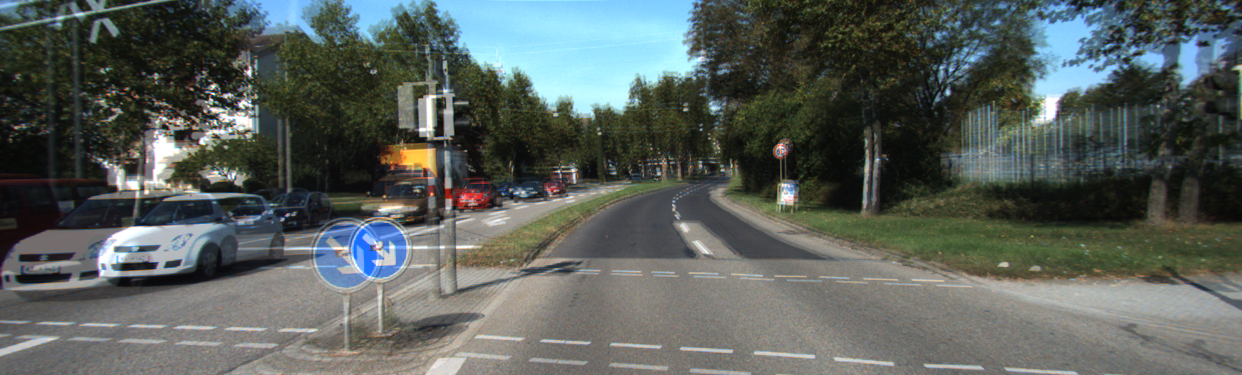}
		\subcaption*{Fused Image}
	\end{minipage}
	\begin{minipage}[b]{0.24\textwidth}
		\includegraphics[scale=0.098]{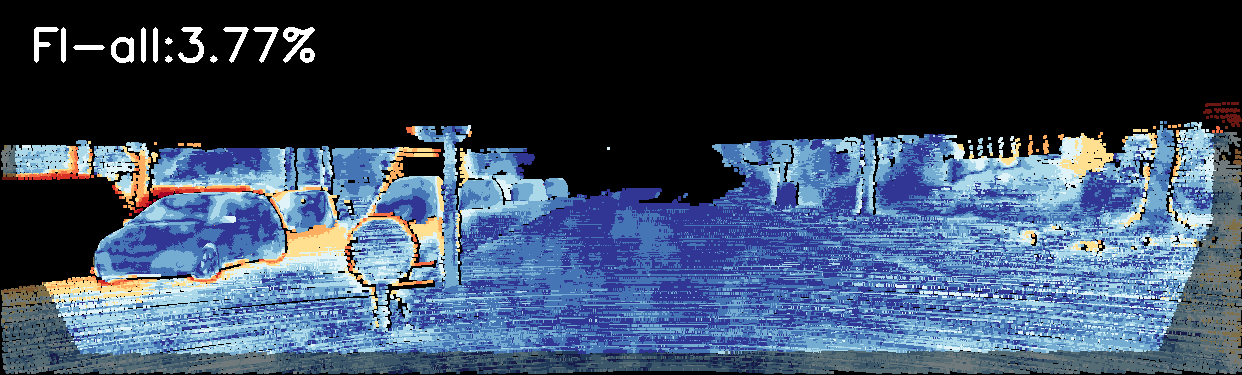}
		\subcaption*{FDFlowNet}
	\end{minipage}
	\begin{minipage}[b]{0.24\textwidth}
		\includegraphics[scale=0.098]{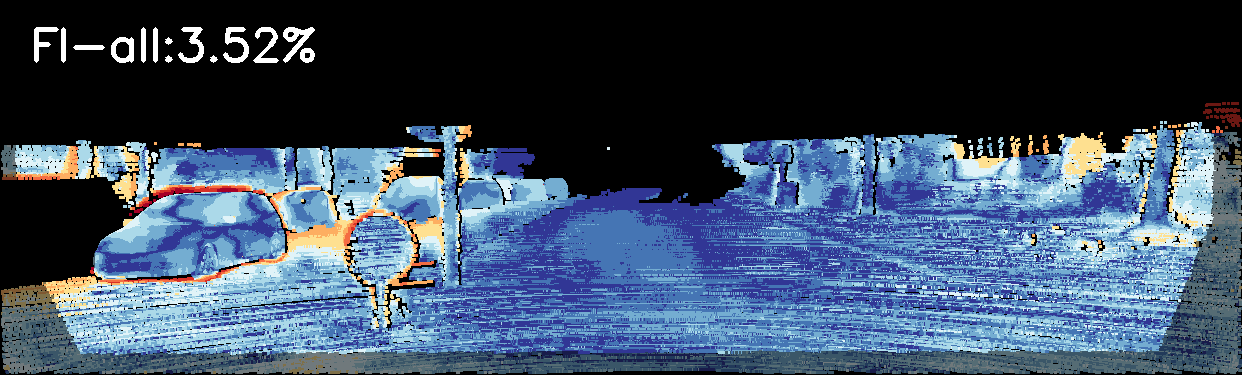}
		\subcaption*{PWC-Net}
	\end{minipage}
	\begin{minipage}[b]{0.24\textwidth}
		\includegraphics[scale=0.098]{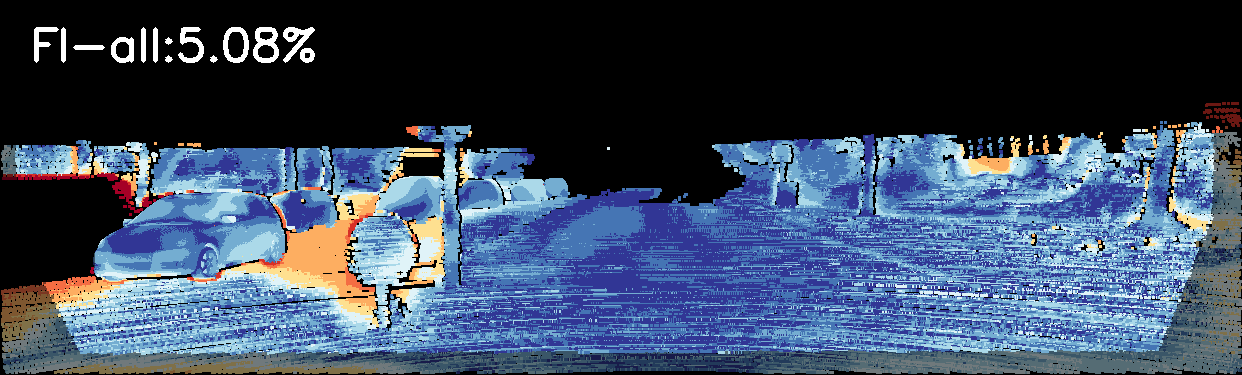}
		\subcaption*{LiteFlowNet}
	\end{minipage}
	\caption{Flow estimation error on the KITTI2015 test set. Red region means large error while blue region denotes small error.}
	\label{fig:4}
\end{figure*}

\subsection{Results}
We test proposed FDFlowNet on benchmarks of Sintel~\cite{Butler:ECCV:2012}, KITTI2012~\cite{Geiger2012CVPR} and KITTI2015~\cite{Menze2015CVPR}. Detailed results are listed in Table~\ref{tab:1}. Default criterion is end point error.

{\noindent \bf MPI Sintel} When fine-tuning on Sinel, we crop $384 \times 768$ patches and remove noise like~\cite{Sun_2018_CVPR, Hui_2018_CVPR}. We adopt batch size of 4 where 2 images from clean part and the others from final part. Training schedule is the same as~\cite{Sun_2018_CVPR}. FDFlowNet outperforms PWC-Net and LiteFlowNet by a large margin on Sintel Clean test set. On the Sintel Final test benchmark, our method is only a little worse than PWC-Net and outperforms all the other approaches. Some visual results are depicted in Fig.~\ref{fig:3}. We can see that FDFlowNet keeps better semantic border in flow fields.

{\noindent \bf KITTI} When fine-tuning on KITTI, we crop $320 \times 896$ patches and reduce magnitude of rotation, zoom and squeeze in data augmentation. Learning schedule is the same as fine-tuning on Sintel. On KITTI2012 test benchmark, FDFlowNet excels all the others. Our method gets the same good results on KITTI2015 test set as LiteFlowNet. Note that LiteFlowNet has lower end point error on the training datasets of Sintel and KITTI which indicates good generalization ability of proposed deep network. Comparison of flow estimation errors among several competitive methods on KITTI2015 test dataset is shown in Fig.~\ref{fig:4}. It can be seen that our model predicts better flow fields in low and repetitive texture areas.

\begin{table}[t]
\small
\renewcommand\arraystretch{1.2}
\begin{center}
\centering
\resizebox{0.48\textwidth}{!}{
\begin{tabular}{c|c|c|c}
	Variants & FDFlowNet & FDFlowNet-U & FDFlowNet-PFC \\
	\hline
	Chairs & {\bf 1.92} & 2.14 & 2.02\\
	\hline
	Sintel Clean & {\bf 3.06} & 3.24 & 3.18\\
	\hline
	Sintel Final & {\bf 4.23} & 4.46 & 4.34\\
\end{tabular}}
\end{center}
\caption
{Ablation study of different variants of FDFlowNet.}
\label{tab:2}
\end{table}

\subsection{Ablation Study}
In this section, we conduct ablation study to evaluate the effectiveness of proposed approaches. Results of end point error are listed in Table~\ref{tab:2}, "FDFlowNet-U" means that U-shape network is removed. "FDFlowNet-PFC" represents substituting sequential connected structure for partial fully connected structure with dilated convolution. All models are trained on FlyingChairs using the same learning schedule and evaluated on Chairs test, Sintel training datasets. Experiments show that U-shape network can provide better feature representation with fused multi-scale information that obtains an obvious improvement. It is about $5.6\%$ improvement on Sintel Clean and about $5.2\%$ improvement on Sintel Final. Partial fully connected structure with dilated convolution also surpass traditional sequential topology.

\subsection{Runtime and Parameters}
It is important for optical flow network to be running fast in real-time and lightweight. This is especially significant in embedded and mobile devices. Here we measure running speed and number of parameters of different optical flow networks as displayed in Table~\ref{tab:3}. Experiments are conducted on a machine equipped with one NVIDIA GTX 1080Ti GPU. We use the PyTorch implement of all networks for fair comparison. Running time is obtained on Sintel resolution $(436 \times 1024)$ averaged over 1000 times. FDFlowNet runs fastest among all the well-behaved models. It is about 2 times faster than PWC-Net and about 3.2 times faster than LiteFlowNet. It also outperforms PWC-Net-small in both running speed and benchmark performance which demonstrates the effectiveness of proposed FDFlowNet and related contributions.

\begin{table}[t]
\small
\renewcommand\arraystretch{1.2}
\centering
\begin{tabular}{c|c|c|c}
	Model & FlowNetC & FlowNet2 & SPyNet \\
	\hline
	parameters (M) & 39.18 & 162.49 & {\bf 1.20} \\
	\hline
	runtime (ms) & 24.6 & 115.7 & 47.4 \vspace{1mm} \\
	Model & PWC-Net & LiteFlowNet & FDFlowNet \\
	\hline
	parameters (M) & 8.75 & 5.37 & 5.79 \\
	\hline
	runtime (ms) & 32.2 & 53.2 & {\bf 16.7} \\
\end{tabular}
\caption{Comparison of model size and running time.}
\label{tab:3}
\end{table}

\section{Conclusion}
This paper has presented a new fast and lightweight deep network for optical flow. By replacing previous pyramid feature with fused feature of U-shape network, our model gets better results on challenging benchmarks. We have proposed a new partial fully connected structure that provides a tradeoff between dense and sequential connected structures in model size, computation cost and network performance. It makes our FDFlowNet run at about 60 fps on Sintel resolution with relatively good performance. We hope our model and related contributions can help vast computer vision applications such as action recognition, video processing and automatic driving.

\bibliographystyle{IEEEbib}
\bibliography{refs}

\end{document}